\documentclass[11pt,a4paper]{article}
\usepackage[hyperref]{naaclhlt2019}
\usepackage{times}
\usepackage{latexsym}

\usepackage{url}

\usepackage{amsmath}
\usepackage{algorithm}
\usepackage{algorithmic}
\usepackage{graphicx}
\usepackage{booktabs}
\usepackage{multirow}
\graphicspath{{./figures/}}

\usepackage[textsize=scriptsize]{todonotes}
\definecolor{blue}{RGB}{0, 93, 170}

\aclfinalcopy 
\def\aclpaperid{2294} 

\let\svthefootnote\thefootnote
\newcommand\blankfootnote[1]{%
  \let\thefootnote\relax\footnotetext{#1}%
  \let\thefootnote\svthefootnote%
}

\title{Sentence Embedding Alignment for Lifelong Relation Extraction
}
\author{
 Hong Wang$^\dagger$,
 Wenhan Xiong$^\dagger$,
 Mo Yu$^{\ddag \ast}$, 
 Xiaoxiao Guo$^{\ddag \ast}$, 
 Shiyu Chang$^\ddag$, 
 William Yang Wang$^\dagger$
\\
 $^\dagger$ University of California, Santa Barbara\\
 $^\ddag$ IBM Research\\
 \{hongwang600, xwhan, william\}@cs.ucsb.edu, yum@us.ibm.com, \{xiaoxiao.guo, shiyu.chang, \}@ibm.com  
 }
\date{}

\begin{document}
\maketitle
\begin{abstract}
  Conventional approaches to relation extraction usually require a fixed set of pre-defined relations. Such requirement is hard to meet in many real applications, especially when new data and relations are emerging incessantly and it is computationally expensive to store all data and re-train the whole model every time new data and relations come in.
  We formulate such a challenging problem as lifelong relation extraction and investigate memory-efficient incremental learning methods without catastrophically forgetting knowledge learned from previous tasks.
  We first investigate a modified version of the stochastic gradient methods with a replay memory, which surprisingly outperforms recent state-of-the-art lifelong learning methods.
  We further propose to improve this approach to alleviate the forgetting problem by anchoring the sentence embedding space.
  Specifically,  we utilize an explicit alignment model to mitigate the sentence embedding distortion of the learned model when training on new data and new relations.
  Experiment results on multiple benchmarks show that our proposed method significantly outperforms the state-of-the-art lifelong learning approaches.\blankfootnote{$^\ast$ Co-mentoring}
  \blankfootnote{$^\_$ Code and dataset can be found in this repository: \url{https://github.com/hongwang600/Lifelong_Relation_Detection}} 
\end{abstract}

\section{Introduction}
The task of relation detection/extraction aims to recognize entity pairs' relationship from given contexts. As an essential component for structured information extraction, it has been widely used in downstream tasks such as automatic knowledge-based completion~\cite{riedel2013relation} and question answering~\cite{P15-1128,DBLP:conf/acl/YuYHSXZ17}. 

Existing relation detection methods always assume a closed set of relations and perform  once-and-for-all training on a fixed dataset. While making the evaluation straightforward, this setting clearly limits the usage of these methods in realistic applications, where new relations keep emerging over time. To build an evolving system which automatically keeps up with the dynamic data, we consider a more practical lifelong learning setting~(also called \emph{continual learning})~\cite{ring1994continual, Thrun1998, thrun2012learning}, where a learning agent learns from a sequence of tasks, where each of them includes a different set of relations. In such scenarios, it is often infeasible to combine the new data with all previous data and re-train the model using the combined dataset, especially when the training set for each task is huge.

To enable efficient learning in such scenarios, recent lifelong learning research \cite{DBLP:journals/corr/KirkpatrickPRVD16,DBLP:conf/nips/Lopez-PazR17} propose to learn the tasks incrementally, while at the same time preventing catastrophic forgetting~\cite{mccloskey1989catastrophic,ratcliff1990connectionist,mcclelland1995there,french1999catastrophic}, i.e., the model abruptly forgets knowledge learned on previous tasks when learning on the new task. Current lifelong learning approaches address such challenge by either preserving the training loss on previously learned tasks (GEM)~\cite{DBLP:conf/nips/Lopez-PazR17}, or selectively dimming the updates on important model parameters (EWC)~\cite{DBLP:journals/corr/KirkpatrickPRVD16}. These methods usually involve adding additional constraints on the model's parameters or the updates of parameters by utilizing stored samples. Despite the effectiveness of these methods on simple image classification tasks, there is little research validating the practical usage of these methods in realistic NLP tasks.  
In fact, when applying these methods to our relation extraction task, we observe that they underperform a simple baseline that updates the model parameters (i.e., learning by SGD) with a mix of stored samples from previous tasks and new samples from the incoming task. We further test this simple baseline on commonly used continual learning benchmarks and get similar observations. 

In this work, we thoroughly investigate two existing continual learning algorithms on the proposed lifelong relation extraction task. We observe that recent lifelong learning methods only operate on the models' parameter space or gradient space, and do not explicitly constraint the feature or embedding space of neural models. As we train the model on the new task, the embedding space might be distorted a lot, and become infeasible for previous tasks.
We argue that 
the embedding space should not be distorted much in order to let the model work consistently on previous tasks. To achieve this, we propose an alignment model that explicitly anchors 
the sentence embeddings derived by the neural model.
Specifically, the alignment model treats the saved data from previous tasks as anchor points and minimizes the distortion of the anchor points in the embedding space in the lifelong relation extraction.  The aligned embedding space is then utilized for relation extraction. Experiment results show that our method outperforms the state-of-the-art significantly in accuracy while remaining efficient.

The main contributions of this work include:

$\bullet$ We introcduce the lifelong relation detection problem and construct lifelong relation detection benchmarks from two datasets with large relation vocabularies: SimpleQuestions \cite{DBLP:journals/corr/BordesUCW15} and FewRel \cite{DBLP:conf/emnlp/HanZYWYLS18}.

$\bullet$ We propose a simple memory replay approach and find that current popular methods such as EWC and GEM underperform this method.

$\bullet$ We propose an alignment model which aims to alleviate the catastrophic forgetting problem by slowing down the fast changes in the embedding space for lifelong learning.

\section{Problem Definition}
\label{sec:definition}

\paragraph{Generic definition of lifelong learning problems}
In lifelong learning, there is a sequence of $K$ tasks $\{\mathcal{T}^{(1)}, \mathcal{T}^{(2)}, \dots ,\mathcal{T}^{(K)}\}$.
Each task $\mathcal{T}^{(k)}$ is a conventional supervised task, with its own label set $L^{(k)}$ and training/validation/testing data ($T^{(k)}_{\textrm{train}}, T^{(k)}_{\textrm{valid}}, T^{(k)}_{\textrm{test}}$), each of which is a set of labeled instances $\{(x^{(k)}, y^{(k)})\}$.
The goal of lifelong learning is to learn a classification model $f$.
At each step $k$, $f$ observes the task $\mathcal{T}^{(k)}$, and optimizes the loss function on its training data with a loss function $\ell(f(x), y)$. 
At the same time, we require the model $f$ learned after step $k$ could still perform well on the previous $k-1$ tasks. That is, we evaluate the model by using the average accuracy of $k$ tasks at each step as $\frac{1}{k}\sum_{j=1}^k acc_{f,j}$.



To make $f$ perform well on the previous tasks, during the lifelong learning process, we usually allow the learner 
to maintain and observe a memory $\mathcal{M}$ of samples from the previous tasks.
Practically, with the growth of the number of tasks, it is difficult to store all the task data\footnote{Even the data can be stored, it is unrealistic to make full usage of the stored data. For example, random sampling from all previous task data (e.g., for the methods in Section \ref{sec:emr}) will become statistically inefficient.}.
Therefore, in lifelong learning research, the learner is usually constrained on the memory size, denoted as a constant $B$. Thus at each step $k$, the learner is allowed to keep training samples from $\{\mathcal{T}^{(j)}|j=1,\dots ,k-1\}$ with size less or equal to $B$.

\paragraph{Lifelong relation detection}
In this paper we introduce a new problem, \emph{lifelong relation detection}.
Relation detection is an important task that aims to detect whether a relation exists between a pair of entities in a paragraph.
In many real-world scenarios, relation detection naturally forms a lifelong learning problem because new relation types emerge as new knowledge is constantly being discovered in various domains. For example, in the Wikidata~\cite{vrandevcic2014wikidata} knowledge graph, the numbers of new items and properties are constantly increasing\footnote{\url{https://www.wikidata.org/wiki/Wikidata:News}}. So
we need to keep collecting data and updating the model over time in order to handle newly added relations.

The problem of lifelong relation detection has the same definition as above with only one difference: 
during prediction time, we hope to know whether an input paragraph contains any relation observed before. Therefore at time $k$, given an input $x$ from task $j'$$<$$k$, instead of predicting an $y \in L^{(j')}$, we predict $y^{(k)}\in \bigcup_{j=1}^k L^{(j)}$. That says, the candidate label set is expanding as the learner observes more tasks, and the difficulty of each previous task is increasing over time as well.



\section{Evaluation Benchmarks for Lifelong Learning}

\subsection{Previous non-NLP Benchmarks}
\paragraph{Lifelong MNIST}
MNIST is a dataset of handwriting ten digits \cite{lecun1998mnist}, where the input for each sample is an image, and the label is the digit the image represents. Two variants 
of the MNIST dataset were proposed for lifelong learning evaluation. One is MNIST \emph{Permutations} \cite{DBLP:journals/corr/KirkpatrickPRVD16}, where a task is created by rearranging pixels according to a fixed permutation. $K$ different permutations are used to generate $K$ tasks. Another variant is MNIST \emph{Rotations}~\cite{DBLP:conf/nips/Lopez-PazR17}, where each task is created by rotating digits by a fixed angle. $K$ angles are chosen for creating $K$ tasks. In our experiments, we follow~\cite{DBLP:conf/nips/Lopez-PazR17} to have $K=20$ tasks for each benchmark.

\paragraph{Lifelong CIFAR}
CIFAR \cite{krizhevsky2009learning} is a dataset used for object recognition, where the input is an image, and the label is the object the image contains. Lifelong CIFAR100 \cite{DBLP:conf/cvpr/RebuffiKSL17} is a variant of CIFAR-100 (CIFAR with 100 classes) by dividing 100 classes into $K$ disjoint subsets. Each task contains samples from $\frac{100}{K}$ classes in one subset. Following~\cite{DBLP:conf/nips/Lopez-PazR17}, we have $K=20$ tasks, where each of them has 5 labels.

\subsection{The Proposed Lifelong Relation Detection Benchmarks}

\paragraph{Lifelong FewRel}
FewRel~\cite{DBLP:conf/emnlp/HanZYWYLS18} is a recently proposed dataset for few-shot relation detection. There are $80$ relations in this dataset.
We choose to create a lifelong benchmark based on FewRel because there are a sufficient number of relation labels.
We extract the sentence-relation pairs from FewRel and build our lifelong FewRel benchmark as follows. Each sample contains a sentence with the relation it refers, and a candidate set of $10$ randomly chosen relations. The model is required to distinguish the right relation from the candidates. We apply K-Means over the averaged word embeddings of the relation names and divide $80$ relations into $10$ disjoint clusters.
This results in $10$ tasks in this benchmark, and each task contains relations from one cluster. 

\paragraph{Lifelong SimpleQuestions}
SimpleQuestions is a KB-QA dataset containing single-relation questions~\cite{DBLP:journals/corr/BordesUCW15}. \cite{DBLP:conf/acl/YuYHSXZ17} created a relation detection dataset from SimpleQuestions that contains samples of question-relation pairs. For each sample, a candidate set of relations is also provided. Similar to lifelong FewRel, we divide relations into $20$ disjoint clusters by using K-Means. This results in $20$ tasks, and each task contains relations from one cluster.

\section{Simple Episodic Memory Replay Algorithm for Lifelong Learning}
\label{sec:emr}
Catastrophic forgetting is one of the biggest obstacles in lifelong learning. The problem is particularly severe in neural network models, because the learned knowledge of previous tasks is stored as network weights, while a slight change of weights when learning on the new task could have an unexpected effect on the behavior of the models on the previous tasks~\cite{french1999catastrophic}.

Currently, the memory-based lifelong learning approaches, which maintain a working memory of training examples from previous tasks, are proved to be one of the best solutions to the catastrophic forgetting problem.
In this section, we first propose a memory-based lifelong learning approach, namely Episodic Memory Replay (EMR), which uses the working memory by sampling stored samples to replay in each iteration of the 
new task learning.
Surprisingly, such a straightforward approach with a clear motivation was never used in previous research. 
We first compare  EMR with the state-of-the-art memory-based algorithm Gradient Episodic Memory (GEM). We also show that the EMR outperforms GEM on many benchmarks, suggesting that it is likely to be among the top-performed lifelong learning algorithms, and it should never be ignored for comparison when developing new lifelong learning algorithms.



\subsection{Episodic Memory Replay (EMR)}
EMR is a modification over stochastic gradient descent algorithms. It replays randomly sampled data from memory while training on a new task, so the knowledge of previous tasks could be retained in the model. 
After training on each task $k$, EMR selects several training examples to store in the memory $\mathcal{M}$, denoted as $\mathcal{M} \bigcap T^{(k)}_{\textrm{train}}$.\footnote{\cite{rebuffi2017icarl} propose to dynamically change the size of memory set for each task during training. The followup work and this paper all use fixed sets, and we will investigate the usage of dynamic sets in future work.}

To handle the scalability, EMR stochastically replays the memory. 
Specifically, when training on task $k$ with each mini-batch $D^{(k)}_{\textrm{train}} \subset T^{(k)}_{\textrm{train}}$, EMR samples from the memory  $\mathcal{M}$
to form a second mini-batch $D^{(k)}_{\textrm{replay}} \subset \mathcal{M}$.
Then two gradient steps are taken on the two mini-batches of $D^{(k)}_{\textrm{train}}$ and $D^{(k)}_{\textrm{replay}}$.
Note that EMR could work with any stochastic gradient optimization algorithm, such as SGD, Adagrad, AdaDelta, and Adam, to optimize the model $f$ with the mixed mini-batches.

We try two variations of $D^{(k)}_{\textrm{replay}}$ sampling: 
first, \emph{task-level sampling}, which samples from one previous task $j$ each time, i.e., $D^{(k)}_{\textrm{replay}} \subset \mathcal{M} \bigcap T^{(j)}_{\textrm{train}}$.
Second, \emph{sample-level sampling}, which samples all over the memory, i.e., $D^{(k)}_{\textrm{replay}} \subset \mathcal{M}$.

The two approaches differ in the task instance sampling probability.
The task-level approach assumes a uniform distribution over tasks, while the sample-level approach has a marginal distribution on tasks that is proportional to the number of their training data in $\mathcal{M}$.\footnote{The two approaches hence favor different evaluation metrics -- the former fits macro averaging better and the latter fits micro averaging better.}
When tasks are balanced like MNIST and CIFAR, or when the stored data in the memory for different tasks are balanced, the two approaches become equivalent.

However, the sample-level strategy could sometimes make the code implementation more difficult: for some lifelong learning benchmarks such as MNIST Rotation, MNIST Permutation, and CIFAR-100 used in \cite{DBLP:conf/nips/Lopez-PazR17}, the tasks could differ from each other in the input or output distribution, leading to different computation graphs for different training examples. 
From our preliminary study, the task-level approach could always give results as good as those of the sample-level approach on our lifelong relation detection benchmarks (see Table \ref{Lifelong_mnist_cifar}) , so in our experiments in Section \ref{sec:exp} we always use the task-level approach.


\subsection{Comparing EMR with State-of-the-art Memory-based Lifelong Algorithm}
In this part, we will first thoroughly introduce a state-of-the-art memory-based lifelong learning algorithm called Gradient Episodic Memory (GEM)~\cite{DBLP:conf/nips/Lopez-PazR17}, and then compare EMR with it in both time complexity and experimental results on several benchmarks.
\paragraph{Gradient Episodic Memory (GEM)}
The key idea of GEM~\cite{DBLP:conf/nips/Lopez-PazR17} is to constrain the new task learning with previous task data stored in memory.
Specifically, it constrains the gradients during training with the following operation.
When training on task $k$, for each mini-batch $D^{(k)}_{\textrm{train}} \subset T^{(k)}_{\textrm{train}}$, it first computes the gradient $g^{(k)}_{\textrm{train}}$ on $D^{(k)}_{\textrm{train}}$,
and the average gradients on the stored data of each previous task $j$, denoted as $g_{\textrm{task}}^{(j)}$.
More concretely, we define
\begin{equation*}
    g_{\textrm{task}}^{(j)}=\frac{ \sum_{i'}\nabla \ell(f(x^{(j)}_{i'}), y^{(j)}_{i'})}{\vert \mathcal{M} \bigcap T^{(j)}_{\textrm{train}} \vert}, \nonumber
\end{equation*}
where $j$$<$$k$, $\ell(\cdot)$ is the loss function, and $(x^{(j)}_{i'}, y^{(j)}_{i'}) \in \mathcal{M} \bigcap T^{(j)}_{\textrm{train}}$, i.e. $(x^{(j)}_{i'}, y^{(j)}_{i'})$ is a training instance in $\mathcal{T}^{(j)}$ that was stored in memory $\mathcal{M}$.
Then the model $f$ is updated along the gradient $\tilde{g}$ that solves the following problem:
\begin{equation*}
\label{GEM_opt}
\begin{aligned}
& \text{min}_{\tilde{g}}
& & ||\tilde{g} - g^{(k)}_{\textrm{train}}||^2 \\
& \text{s.t.}
& & \langle \tilde{g}, g_{\textrm{task}}^{(j)} \rangle \geq 0, \; j = 1, \ldots, k-1.
\end{aligned}
\end{equation*}
$\tilde{g}$ is the closest gradient to the gradient on the current training mini-batch, $g^{(k)}_{\textrm{train}}$, without decreasing performance on previous tasks much since the angle between $\tilde{g}$ and $g_{\textrm{task}}^{(j)}$ is smaller than $90^{\circ}$. 

\paragraph{Time Complexity}
One difference between EMR and GEM is that EMR deals with unconstrained optimization and does not require the gradient projection, i.e., solving $\tilde{g}$. But since the model $f$ is deep networks, empirically the time complexity is mainly dominated by the computation of forward and backward passes. We analyze the time complexity as below:

In task $k$, suppose the mini-batch size is $\vert D \vert$ and the memory replay size is $m$, our EMR takes $\vert D \vert+m$ forward/backward passes in each training batch. Note that $m$ is a fixed number and set to be equal to the number of instances stored for each previous task in our experiments.
While for GEM, it needs to compute the gradient of all the data stored in the memory $\mathcal{M}$, thus $\vert D \vert + \vert \mathcal{M} \vert$ forward/backward passes are taken. Its complexity is largely dominated by the size $\vert \mathcal{M} \vert$ (upper bounded by the budget $B$).
When the budget $B$ is large, with the number of previous tasks increases, $\mathcal{M}$ grows linearly, and GEM will become infeasible.



\paragraph{Superior Empirical Results of EMR}

\begin{table}[t]
\centering
\resizebox{0.48\textwidth}{!}{
 \begin{tabular}{c | c c | c} 
 \toprule
  \multirow{2}{*}{Task} & 
 \multicolumn{2}{c|}{EMR} &  \multirow{2}{*}{GEM}\\\cline{2-3}
 &sample & task  \\
 \midrule
 MNIST Rotation & --&0.828&\bf 0.860 \\
 MNIST Permutation &-- &0.824&\bf 0.826 \\
 CIFAR-100 &--& \bf 0.675&\bf 0.675\\
 \hline
 FewRel & 0.606 & \bf 0.620 & 0.598\\
 SimpleQuestions & 0.804 & \bf 0.808 & 0.796\\
 \bottomrule
\end{tabular}
}
\vspace{-0.1in}
\caption{
The average accuracy across all the tasks at last time step for EMR and GEM on both non-NLP and our lifelong relation detection benchmarks. For the experiments on MNIST and CIFAR, we follow the setting in ~\cite{DBLP:conf/nips/Lopez-PazR17} (see Appendix \ref{appendix:mnist_setting} for details). For the experiments on FewRel and SimpleQuestions, we use the same setting in Section \ref{sec:exp}.
We only implement task-level EMR for MNIST and CIFAR because of the relatively easy implementation.}
\vspace{-0.1in}
\label{Lifelong_mnist_cifar}
\end{table}


The EMR algorithm is much simpler compared to the GEM.
However, one interesting finding of this paper is that the state-of-the-art GEM is unnecessarily more complex and more inefficient, because EMR, a simple stochastic gradient method with memory replay, outperforms it on several benchmarks.

The results are shown in Table \ref{Lifelong_mnist_cifar}.
The numbers are the average accuracy, i.e. $\frac{1}{k}\sum_{j=1}^k acc_{f,j}$, at last time step.
For both algorithms, the training data is randomly sampled to store in the memory, following~\cite{DBLP:conf/nips/Lopez-PazR17}.
On lifelong relation detection, the EMR outperforms GEM on both of our created benchmarks. To further show its generalizability, we apply the EMR to previous lifelong MNIST and CIFAR benchmarks and compare to the results in \cite{DBLP:conf/nips/Lopez-PazR17} with all the hyperparameters set as the same.
Still, EMR performs similarly to GEM except for the MNIST Rotation benchmark.\footnote{Even on MNIST Rotation, it has achieved a competitive result, since the conventional training on shuffled data from all the tasks in this benchmark gives $\sim0.83$ according to~\cite{DBLP:conf/nips/Lopez-PazR17}.}

From the above results, we learned the lesson that previous lifelong learning approaches actually fail to show improvement compared to doing memory replay in a stochastic manner.
We hypothesise that GEM performs worse when there is positive transfer among tasks, making the gradient projection an inefficient way to use gradients computed from memory data.
Therefore, in the next section, we start with the basic EMR and focus on more efficient usage of the historical data. 





\section{Embedding Aligned EMR (EA-EMR)}

Based on our basic EMR, this section proposes our solution to lifelong relation detection.
We improve the basic EMR with two motivations: (1) previous lifelong learning approaches work on the parameter space. 
{However, the number of parameters in a deep network is usually huge. Also, deep networks are highly non-linear models, and the parameter dimensions have complex interactions, making the Euclidean space of parameters not a proper delegate of model behavior~\cite{french1999catastrophic}. That is, a slight change in parameter space could affect the model prediction unexpectedly.
The above two reasons make it hard to maintain deep network behaviors on previous tasks with constraints or Fisher information.
Therefore, we propose to alleviate catastrophic forgetting in the hidden space (i.e., the sentence embedding space).}
(2) for each task, we want to select the most informative samples to store in the memory, instead of random sampling like in \cite{DBLP:conf/nips/Lopez-PazR17}. Therefore the budget of memory can be better utilized.

\subsection{Embedding Alignment for Lifelong Learning}

This section introduces our approach which performs lifelong learning in the embedding space, i.e., the Embedding Aligned EMR (EA-EMR).


In EA-EMR, for each task $k$, besides storing the original training data $(x^{(k)},y^{(k)})$ in the memory $\mathcal{M}$, we also store their embeddings. 
In the future after a new task is trained, the model parameters are changed thus the embeddings for the same $(x^{(k)},y^{(k)})$ would be different. 
Intuitively, a lifelong learning algorithm should allow such  parameter changes but ensure the changes do not distort the previous embedding spaces too much.

Our EA-EMR alleviates the distortion of embedding space with the following idea: if the embedding spaces at different steps are not distorted much, there should exist a simple enough transformation $a$ (e.g., a linear transformation in our case) that could transform the newly learned embeddings to the original embedding space, without much performance degeneration on the stored instances. So we propose to add a transformation $a$ on the top of the original embedding and learn the basic model $f$ and the transformation $a$ automatically. Specifically, at the $k$-th task, we start with the model $f^{(k-1)}$, and the transformation $a^{(k-1)}$, that trained on the previous $k-1$ tasks. 
We want to learn the basic model $f$ and the transformation $a$ such that the performance on the new task and stored instances are optimized without distorting the previous embedding spaces much.
\begin{equation*}
\label{objective_function}
\small
\begin{aligned}
    &\min_{f(\cdot), a(\cdot)} \sum_{(x,y)\in D_{\textrm{train}}^{(k)}} \ell(a(f(x)), y) \nonumber  + \\
    & \sum_{(x,y) \in D_{\textrm{replay}}^{(k)}} \left( \ell(a(f(x)), y) + \Vert a(f(x)) -a^{(k-1)} f^{(k-1)}(x) \Vert^2 \right) \nonumber
\end{aligned}
\end{equation*}

We propose to minimize the above objective through two steps. In the first step, we optimize the basic model $f$ by:
\begin{equation}
\label{step_1}
\small
\begin{aligned}
    \min_{f(\cdot)} &\sum_{(x,y)\in D_{\textrm{train}}^{(k)} \bigcup D_{\textrm{replay}}^{(k)} } \ell \left(a^{(k-1)}f(x), y \right) \nonumber
\end{aligned}
\end{equation}
This step mainly focuses on learning the new task without performance drop on the stored samples.

In the second step, we optimize $a$ to keep the embedding space of the current task and restore the previous embedding space of stored samples: 
\begin{equation*}
\label{step_2}
\small
\begin{aligned}
    \min_{a(\cdot)} &\sum_{(x,y) \in D_{\textrm{train}}^{(k)}}\Vert a(f(x)) - a^{(k-1)}f(x) \Vert^2 \\
    &+ \sum_{(x,y) \in D_{\textrm{replay}}^{(k)}}\Vert a(f(x)) - a^{(k-1)}f^{(k-1)}(x) \Vert^2  
\end{aligned}
\end{equation*}

\begin{figure}[!t]
\centering
\includegraphics[width=6cm]{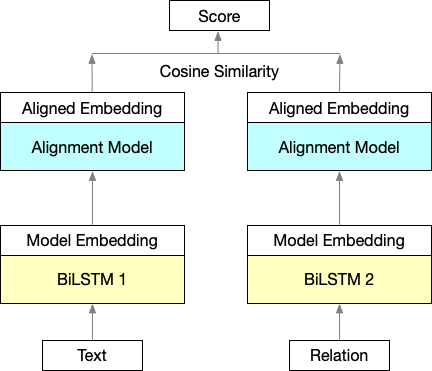}
\vspace{-0.1in}
\caption{This figure shows how we add the alignment model (a linear model in our case) on the basic relation detection model, where two BiLSTMs are used to encode the text and relation, and cosine similarity between their embeddings are computed as the score.}
\label{model_png}
\vspace{-0.1in}
\end{figure}

\paragraph{Embedding Alignment on Relation Detection Model}
We introduce how to add embedding alignment to relation detection models. The basic model we use is a ranking model that is similar to \text{HR-BiLSTM} \cite{DBLP:conf/acl/YuYHSXZ17}.
Two \text{BiLSTMs} \cite{DBLP:journals/neco/HochreiterS97} are used to encode the sentence and relation respectively given their GloVe word embedding \cite{pennington2014glove}. Cosine similarity between the sentence and relation embedding is computed as the score. Relation with maximum score is predicted by the model for the sentence. Ranking loss is used to train the model\footnote{Though the basic model is simple, it achieves reasonable results on the two datasets when training with all the data, i.e., $0.837$ on FewRel and $0.927$ on SimpleQuestions.}.
This base model is our model $f$, which is trained on a new task $k$ at each step and results in an updated model $f^{(k)}$.
Our proposed approach (Figure \ref{model_png}) inserts an alignment model $a$ to explicitly align to embedding space for stored instances and maintain the embedding space of the current task. Note that the label $y$ (the relation here) also has embedding, so it needs to pass through the alignment model $a$ as well. 

\subsection{Selective Storing Samples in Memory}

When the budget of memory is relatively smaller, how to select previous samples will greatly affect the performance. 
%
Ideally, in order to make the memory best represents a previous task, we hope to choose diverse samples that best approximate the distribution of task data.
However, distribution approximation itself is a hard problem and will be inefficient due to its combinatorial optimization nature. 
Therefore, many recent works such as GEM ignore this step and randomly select samples from each task to store in the memory.

\citet{rebuffi2017icarl} proposed to select exemplars that best approximate the mean of the distribution. This simplest distribution approximation does not give an improvement in our experiments because of the huge information loss.
Therefore, we propose a better approach of sample selection by clustering over the embedding space from the model.
The embedding after alignment model is used to represent the input because the model makes prediction based on that.
Then we apply K-Means (the number of clusters equals the budget given to the specific task) to cluster all the samples of the task. For each cluster, we select the sample closest to the centroid to store in the memory.

We leave more advanced approaches of representative sample selection and their empirical comparison to future work.

\section{Experiments}
\label{sec:exp}


\subsection{Experimental Setting}

\begin{figure*}[!th]
\centering
\begin{minipage}{0.48\linewidth}\centering
\includegraphics[width=7cm]{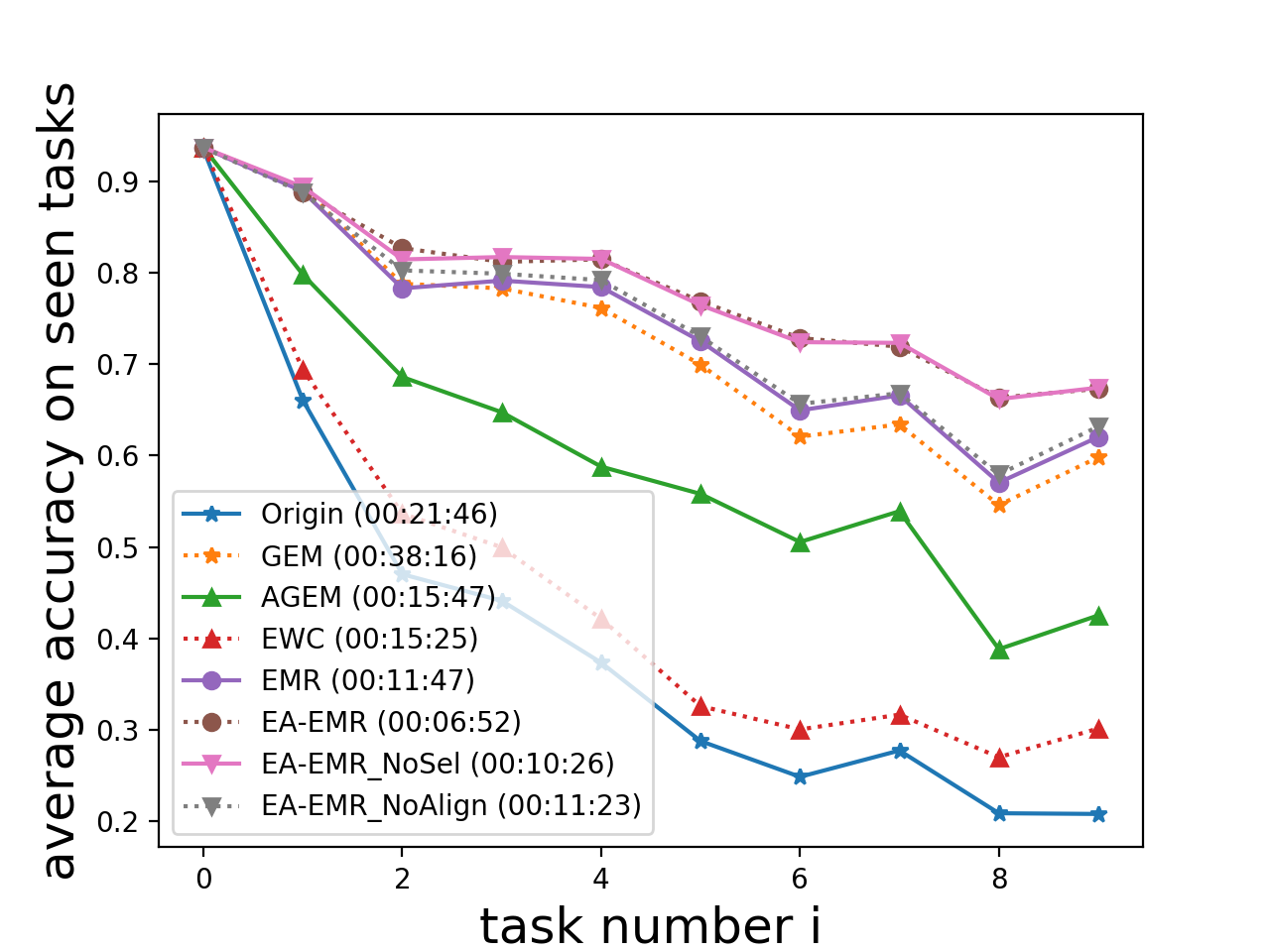}\\
(a) FewRel
\end{minipage}
\begin{minipage}{0.48\linewidth}\centering
\includegraphics[width=7cm]{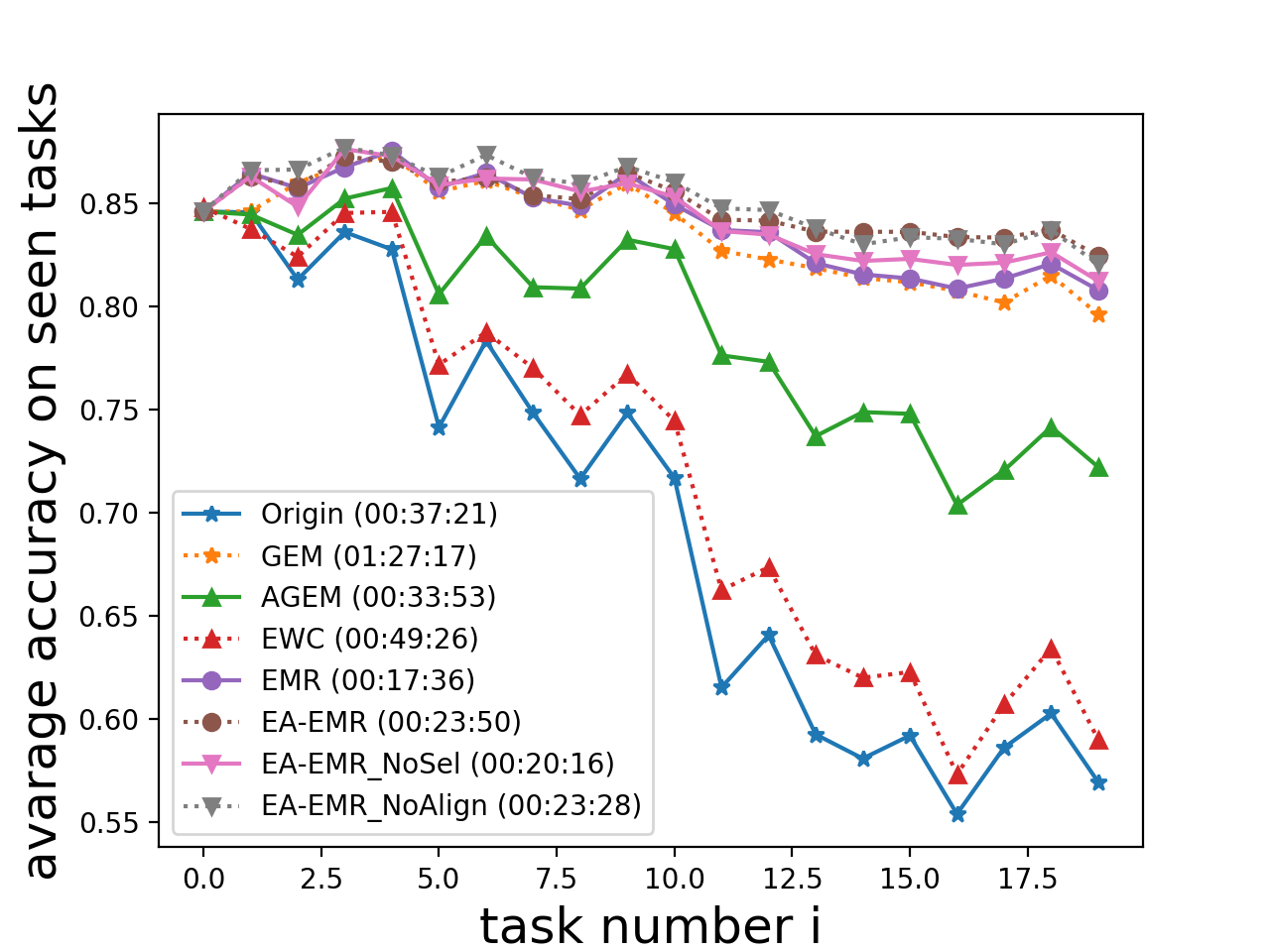}\\
(b) SimpleQuestions
\end{minipage}
\vspace{-0.1in}
\caption{This figure shows the average accuracy of all the observed tasks on the benchmarks of lifelong FewRel and lifelong SimpleQuestions during the lifelong learning process. 
The average performance of $5$ runs is reported, and the \textbf{average running time} is shown in the brackets.}
\label{simpleQustionAvgAcc}
\vspace{-0.1in}
\end{figure*}

We conduct experiments on our lifelong benchmarks: lifelong SimpleQuestions \cite{DBLP:journals/corr/BordesUCW15} and lifelong FewRel \cite{DBLP:conf/emnlp/HanZYWYLS18} to compare our proposed methods
\text{EA-EMR}, EA-EMR without Selection (EA-EMR\_NoSel), EA-EMR without Alignment (EA-EMR\_noAlign), and EMR with the following baselines.

    $\bullet$ \textbf{Origin}, which simply trains on new tasks based on the previous model.
    
    $\bullet$ \textbf{EWC}~\cite{DBLP:journals/corr/KirkpatrickPRVD16}, which slows down updates on important parameters by adding $L_2$ regularization of parameter changes to the loss.
    
    $\bullet$ \textbf{GEM}~\cite{DBLP:conf/nips/Lopez-PazR17}, which projects the gradient to benefit all the tasks so far by keeping a constraint for each previous task.
    
    $\bullet$ \textbf{AGEM}~\cite{anonymous2019efficient}, which only uses one constraint that the projected gradient should decrease the average loss on previous tasks.


 On both FewRel and SimpleQuestions, the epoch to train on each task is set to be $3$. Learning rate for the basic model is set to be $0.001$. The hidden size of LSTM is set to be $200$. The batch size is set to be $50$. For each sample in the memory, $10$ candidate relations is randomly chosen from all observed relations to alleviate the problem that new relations are emerging incessantly.
 
 Parameters for our model and baselines are set as follows. For EA-EMR and EA-EMR\_NoSel, when training the alignment model, the learning rate is set to be 0.0001, and the training epoch is set to be 20 and 10 for FewRel and SimpleQuestions respectively. For AGEM, $100$ samples are randomly chosen from all the previous tasks to form a constraint. For EWC, we set the balancing parameter $\alpha=100$. For GEM and EMR related methods, the batch size of each task is set to be $50$.


\subsection{Lifelong Relation Detection Results}
\paragraph{Evaluation Metrics}\
We use two metrics to evaluate the performance of the model:

\noindent $\bullet$ Average performance on all seen tasks after time step $k$, which highlights the catastrophic problem:
\vspace{-0.1in}
$$
\text{ACC}_{\text{avg}} = \frac{1}{k}\sum_{i=1}^k acc_{f, i}
\vspace{-0.1in}
$$

\noindent $\bullet$ Accuracy on whole testing task:
\vspace{-0.1in}
$$
\text{ACC}_{\text{whole}} = acc_{f, D_{\text{test}}}
\vspace{-0.1in}
$$

\begin{table}[!t]
\small
\centering
\resizebox{0.48\textwidth}{!}{
\begin{tabular}{l | c c |c c}
\toprule
    \multirow{2}{*}{Method} & \multicolumn{2}{c|}{FewRel} & \multicolumn{2}{c}{SimpleQuestions}\\
    &Whole & Avg & Whole & Avg\\
    \midrule
    Origin & 0.189 & 0.208& 0.632 &0.569\\
    \hline
    \multicolumn{5}{c}{\it Baselines}\\
    GEM & 0.492 &0.598& 0.841 &0.796\\
    AGEM & 0.361 &0.425& 0.776 &0.722 \\
    EWC & 0.271 &0.302& 0.672 &0.590 \\
    \hline
    \multicolumn{5}{c}{\it Ours}\\
    Full EA-EMR & \bf 0.566 &0.673& \bf 0.878 &\bf 0.824 \\
    \quad w/o Selection & 0.564 &\bf 0.674& 0.857 &0.812 \\
    \quad w/o Alignment & 0.526 &0.632& 0.869 &0.820\\
    \quad w/o Alignment but keep& \multirow{2}{*}{0.545} & \multirow{2}{*}{0.655} & \multirow{2}{*}{0.871} & \multirow{2}{*}{0.813}\\
    \quad \quad \quad the architecture & & & & \\
    EMR Only & 0.510 &0.620& 0.852 &0.808\\
\bottomrule
\end{tabular}
}
\vspace{-0.1in}
\caption{This table shows the accuracy on the whole test data ("Whole" column), and average accuracy on all observed tasks ("Avg" column) after the last time step.
The average performance of $5$ runs are listed here and the best result on each dataset is marked in bold.}
\vspace{-0.1in}
\label{SimpleQuestionsLastTaskResult}
\end{table}

\paragraph{Results on FewRel and SimpleQuestions}
We run each experiment $5$ times independently by shuffling sequence of tasks, and the average performance is reported.
The average accuracy over all observed tasks during the whole lifelong learning process is presented in Figure \ref{simpleQustionAvgAcc}, and the accuracy on the whole test data during the process is shown in Appendix \ref{append:result_over_time}. 
We also list the result at last step in Table \ref{SimpleQuestionsLastTaskResult}. From the results, we can see that EWC and GEM are better than the Origin baseline on both two datasets, which indicates that they are able to reduce the catastrophic forgetting problem. However, our EA-EMR perform significantly better than these previous state-of-the-arts. The proposed EMR method itself achieves better results than all baselines on both datasets. The ablation study shows that both the \emph{selection} and the \emph{alignment} modules help on both tasks.


\paragraph{The Effect of Embedding Alignment}
To investigate the effect of our embedding alignment approach, we conduct two ablation studies as below:
First, we remove both the alignment loss in equation \ref{step_2}, as well as the alignment module $a$, which results in significant drop on most of the cases (the line ``w/o Alignment'' in Table \ref{SimpleQuestionsLastTaskResult}).
Second, to make sure that our good results do not come from introducing a deeper model with the module $a$, we propose to only remove the embedding alignment loss, but keep everything else unchanged.
That means, we still keep the module $a$ and the training steps, with the only change on replacing the loss in step $2$ with the one in step $1$ (the line ``w/o Alignment but keep the architecture'' in Table \ref{SimpleQuestionsLastTaskResult}).
We can see that this decreases the performance a lot. 
The above results indicate that by explicitly doing embedding alignment, the performance of the model can be improved by alleviating the distortion of previous embedding space.

\paragraph{Comparison of Different Sample Selection Strategies}

Here we compare different selection methods on lifelong FewRel and SimpleQuestions. EMR Only randomly choose samples. \cite{rebuffi2017icarl} propose to choose samples that can best approximate the mean of the distribution.
We compare their sampling strategy (denoted as iCaRL) with our proposed method (K-Means) which encourages to choose diverse samples by choosing the central sample of the cluster in the embedding space. From the results in Table \ref{iCaRLandEMR}, we can see that our method outperforms iCaRL and the random baseline. While iCaRL is not significantly different from the random baseline.

\begin{table}[!t]
\small
\centering
\begin{tabular}{l | c c |c c}
\toprule
    \multirow{2}{*}{Method} & \multicolumn{2}{c|}{FewRel} & \multicolumn{2}{c}{SimpleQuestions}\\
    &Whole & Avg & Whole & Avg\\
    \midrule
    EMR Only & 0.510 &0.620& 0.852 &0.808\\
    + K-Means & \bf 0.526 &\bf 0.632& \bf 0.869 &\bf 0.820\\
    + iCaRL & 0.501 & 0.615 & 0.854 & 0.806\\
\bottomrule
\end{tabular}
\vspace{-0.1in}
\caption{Comparison of different methods to select data for EMR.
The accuracy on the whole test data ("Whole" column), and average accuracy on all observed tasks ("Avg" column) is reported. We run each method $5$ times, and give their average results.}
\vspace{-0.1in}
\label{iCaRLandEMR}
\end{table}

\section{Related Work}

\paragraph{Lifelong Learning without Catastrophic Forgetting}
Recent lifelong learning research mainly focuses on overcoming the \emph{catastrophic forgetting} phenomenon \cite{french1999catastrophic,mccloskey1989catastrophic,mcclelland1995there,ratcliff1990connectionist}, i.e., knowledge of previous tasks is abruptly forgotten when learning on a new task.

Existing research mainly follow two directions: 
the first one is \emph{memory-based approach} \cite{DBLP:conf/nips/Lopez-PazR17,anonymous2019efficient}, which saves some previous samples and optimizes a new task with a forgetting cost defined on the saved samples.
These methods have shown strength in alleviating catastrophic forgetting, but the computational cost grows rapidly with the number of previous tasks.
The second direction is to \emph{consolidate parameters that are important to previous tasks} \cite{DBLP:journals/corr/KirkpatrickPRVD16,DBLP:journals/corr/abs-1802-02950,Ritter2018OnlineSL,DBLP:conf/icml/ZenkePG17}. For example, Elastic Weight Consolidation (EWC) \cite{DBLP:journals/corr/KirkpatrickPRVD16} slows down learning on weights that are important to previous tasks. These methods usually do not need to save any previous data and only train on each task once. But their abilities to overcome catastrophic forgetting are limited.

\paragraph{Lifelong Learning with Dynamic Model Architecture}
There is another related direction on dynamically changing the model structure (i.e., adding new modules) in order to learn the new task without interfering learned knowledge for previous tasks, such as \cite{Xiao:2014:EIL:2647868.2654926,DBLP:journals/corr/RusuRDSKKPH16,DBLP:journals/corr/FernandoBBZHRPW17}.
These approaches could successfully prevent forgetting. However, they do not suit many lifelong settings in NLP. First, it cannot benefit from the positive transfer between tasks. Second, the size of the model grows dramatically with the number of observed tasks,
which makes it infeasible for real-world problems where there are a lot of tasks.

\paragraph{Remark}
It is worth noting that the term lifelong learning is also widely used in~\cite{P15-2123,N15-2018,D16-1022,P17-2023}, 
which mainly focus on how to represent, reserve and extract knowledge of previous tasks. These works belong to a research direction different from lifelong learning without catastrophic forgetting. 

\section{Conclusion}

In this paper, we introduce lifelong learning into relation detection, and find that two state-of-the-art lifelong learning algorithms, GEM and EWC, are outperformed by a simple memory replay method EMR on many benchmarks. Based on EMR, we further propose to use embedding alignment to alleviate the problem of embedding space distortion, which we think is one reason that causes catastrophic forgetting. Also, we propose to choose diverse samples to store in the memory by conducting K-Means in the model embedding space. Experiments verify that our proposed methods significantly outperform other baselines.

\section*{Acknowledgement}
This research was supported in part by a UCSB Chancellor's Fellowship, an IBM Faculty Award, and a DARPA Grant D18AP00044 funded under the DARPA YFA program. The authors are solely responsible for the contents of the paper, and the opinions expressed in this publication do not reflect those of the funding agencies.

\bibliography{naaclhlt2019}
\bibliographystyle{acl_natbib}

\appendix
\clearpage
\onecolumn

\section{Appendix}
\subsection{Performance on the whole test data over time}
\label{append:result_over_time}
\begin{figure*}[!ht]
\centering
\begin{minipage}{0.48\linewidth}\centering
\includegraphics[width=7cm]{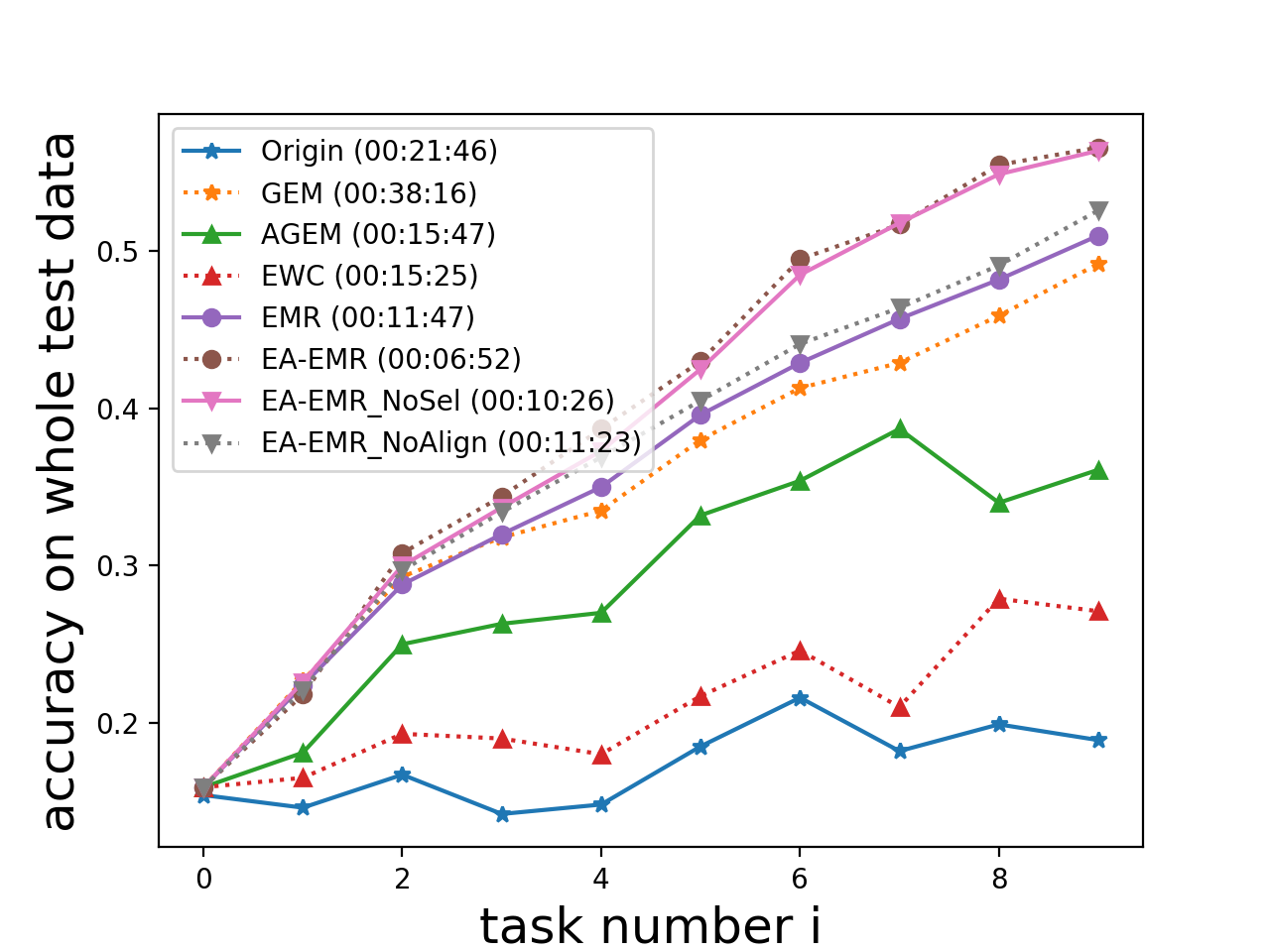}\\
(a) FewRel
\end{minipage}
\begin{minipage}{0.48\linewidth}\centering
\includegraphics[width=7cm]{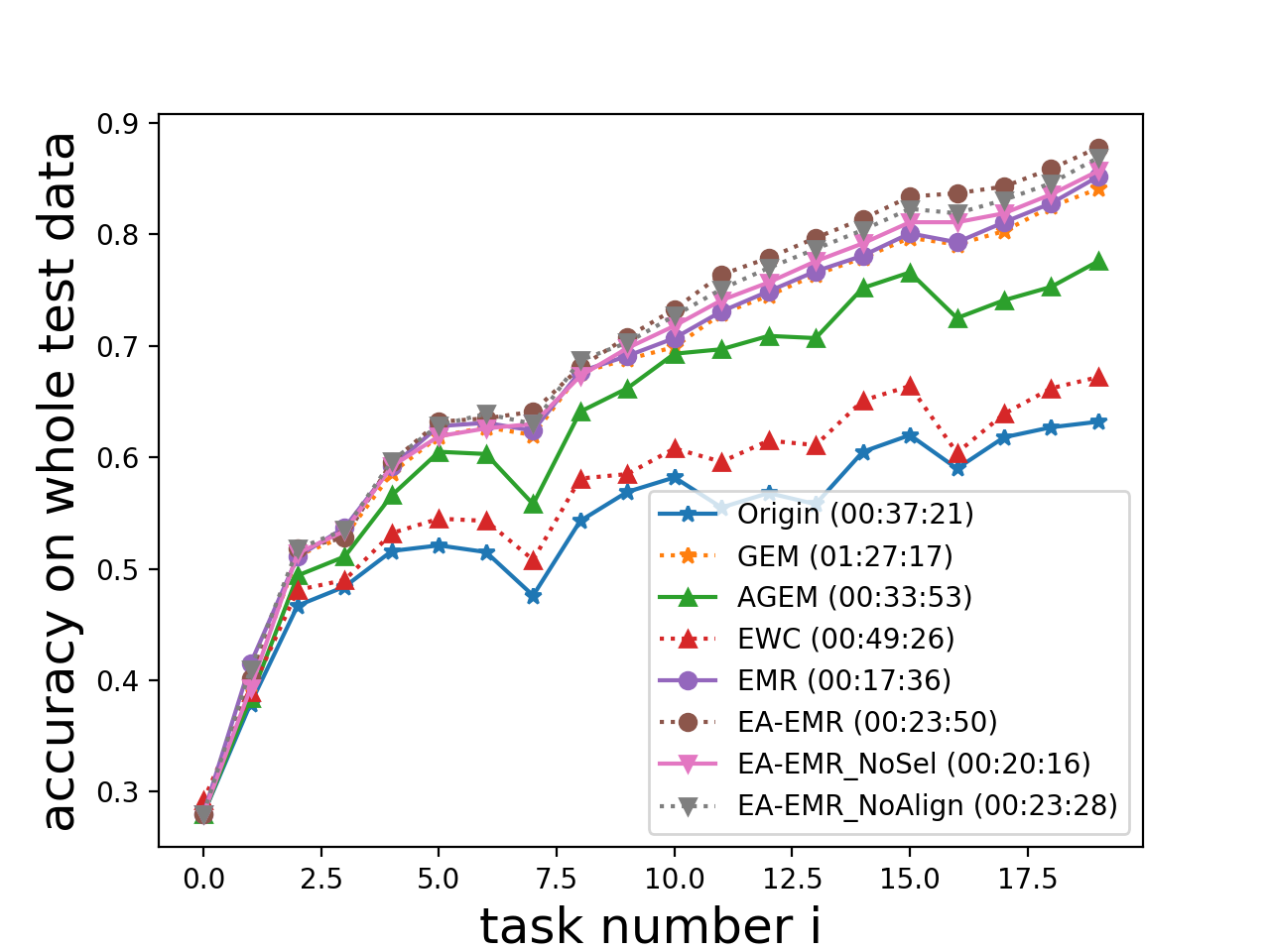}\\
(b) SimpleQuestions
\end{minipage}

\caption{This figure shows the accuracy on the whole test data on the benchmark of lifelong FewRel and lifelong SimpleQuestions during the lifelong learning process. 
The average performance of $5$ runs is reported, and the \textbf{average running time} is shown in the brackets.}
\label{simpleQustionWholeTest}
\vspace{-0.1in}
\end{figure*}

\subsection{Experiment setting for MNIST and CIFAR}
\label{appendix:mnist_setting}
Following the setting in ~\cite{DBLP:conf/nips/Lopez-PazR17}, the size of memory for each task is set to be $256$. The learning rate is set to be $0.1$. The epoch for training the model on each task is set to be $1$. Plain SGD and minibatch of $10$ samples are used. For the MNIST dataset, each task has $1000$ samples of $10$ classes. For the CIFAR dataset, each task has $2500$ samples of $5$ classes.

\end{document}